\documentclass{article}

\usepackage{PRIMEarxiv}

\usepackage[utf8]{inputenc} % allow utf-8 input
\usepackage[T1]{fontenc}    % use 8-bit T1 fonts
\usepackage{hyperref}       % hyperlinks
\usepackage{url}            % simple URL typesetting
\usepackage{booktabs}       % professional-quality tables
\usepackage{amsfonts}       % blackboard math symbols
\usepackage{nicefrac}       % compact symbols for 1/2, etc.
\usepackage{microtype}      % microtypography
\usepackage{lipsum}
\usepackage{fancyhdr}       % header
\usepackage{graphicx}       % graphics
\graphicspath{{media/}}     % organize your images and other figures under media/ folder
\usepackage{pgfplots}

\pgfplotsset{compat=1.7}

\title{Performance Comparison of Simple Transformer and Res-CNN-BiLSTM for Cyberbullying Classification}

\author{
  Raunak Joshi \\
  Research Scholar \\
  University of Mumbai \\
  Mumbai, MH, India \\
  \texttt{raunakjoshi.m@gmail.com} \\
   \And
  Abhishek Gupta \\
  Engineer \\
  University of Mumbai \\
  Mumbai, MH, India \\
  \texttt{abhishekgupta20001@gmail.com } \\
}

\begin{document}
\maketitle

\begin{abstract}
The task of text classification using Bidirectional based LSTM architectures is computationally expensive and time consuming to train. For this, transformers were discovered which effectively give good performance as compared to the traditional deep learning architectures. In this paper we present a performance based comparison between simple transformer based network and Res-CNN-BiLSTM based network for cyberbullying text classification problem. The results obtained show that transformer we trained with 0.65 million parameters has significantly being able to beat the performance of Res-CNN-BiLSTM with 48.82 million parameters for faster training speeds and more generalized metrics. The paper also compares the 1-dimensional character level embedding network and 100-dimensional glove embedding network with transformer.
\end{abstract}

% keywords can be removed

\section{Introduction}
The Natural Language Processing abbreviated as NLP \cite{reshamwala2013review,young2018recent,8629198} has shown a lot of advancement using Deep Learning \cite{lecun2015deep,goodfellow2016deep} system in recent years. The tasks of NLP are varied in terms of applications, viz. Sentiment Analysis \cite{catal2009systematic}, Machine Translation \cite{tan2020neural}, Named Entity Recognition \cite{8629225,li2020survey,roy2021recent} etcetera. Machine Learning \cite{sah2020machine,dhall2020machine} significantly played in important role earlier in the area of NLP before use of Deep Learning was normalized like modern state-of-the-art models. The NLP performs text preprocessing and generates word representations \cite{naseem2021comprehensive} which are the root of the language vocabulary. Then the word representations that hold the meaning of sentences are given to Machine Learning algorithms for task of prognostication. Initially the word representations like bag-of-words \cite{zhang2010understanding} and TF-IDF \cite{kim2019research,Ramos2003UsingTT,tambe2022effects} were used which lacked in providing the semantics in deeper context. These were used for generating moderate level word representations and how the words are linked with each other but were inefficient for providing the literal meaning of a sentence. This was later replaced with advanced form of word representation mechanisms like word embeddings \cite{almeida2019word} and Word2Vec\cite{mikolov2013efficient}. These consisted of word representations on very higher level where sparse matrix representations are used with one hot encoding \cite{cerda2018similarity} system for linkage of the words. These embeddings were high dimensional and used deep learning procedures in itself for better understanding of the semantics of language. The representations held the association of word linkage on multi-dimensional level. Later these representations were replaced by Glove \cite{Pennington2014GloVeGV} which is considered as global vectors for word representations which used context of latent semantic analysis \cite{Landauer1998AnIT} at a very different level which falls under the category of topic modeling \cite{Mahmood2013LiteratureSO}. These embeddings were related to word level embeddings and later the concept of character level embedding \cite{zhang2015character} also started to gain popularity which states that learning characters and making a random word out of context can also generate some semantics of the words respectively. This uses a 1-dimensional character level embedding \cite{8682194} for capturing the character level representations. These all embeddings constitute of the major preprocessing phase and prognostication phase is not limited to machine learning algorithms viz. logistic regression \cite{Cramer2002TheOO}, support vector machines \cite{708428,zhang2012support}, k neighbors \cite{cunningham2021k}, bagging ensemble methods \cite{breiman1996bagging,breiman2001random,kanvinde2022binary}, boosting ensemble methods \cite{freund1996experiments,10.5555/1624312.1624417,schapire2003boosting,gupta2021succinct}, discriminant analysis \cite{Ramayah2010DiscriminantA,gupta2022discriminant}, stacked generalization \cite{Wolpert1992StackedG,nair2022combining} and much more. Deep Learning was leveraged for the same reason where the use of Recurrent Neural Network \cite{cho2014learning} became evident for sequence learning procedures. This is prominently used for time series forecasting and NLP tasks. It can be also used for audio processing tasks too, basically wherever there is a data in sequential manner. The use of it was good but it ran into loss of information and vanishing gradient problem \cite{10.1142/S0218488598000094} while backward propagation \cite{rumelhart1986learning}. For this Long Short Term Memory abbreviated as LSTM \cite{hochreiter1997long,olah2015understanding,staudemeyer2019understanding} was developed which held the necessary amount of information required for the sequence. This actually resolved the vanishing gradient problem but was quite slow to train and computationally expensive back then. Later it was discovered that learning the sequences ahead in time is also possible using Bidirectional \cite{650093} LSTM architecture which links back the learning blocks in forward propagation which turned out to be efficient way more than expected yet turned out to be computationally expensive and time consuming. For this reason the Transformers \cite{vaswani2017attention} were introduced which were one of the most promising learning systems developed in the area of NLP. These were extremely cost efficient and effective that focus on self-attention mechanism which will be discussed in further sections of the paper. The use of transformers ensure much promising results which we are presenting in this paper for cyberbullying \cite{9378065} application. The use of transformers and how the difference can be made for our custom trained combination of traditional techniques condensed neural network, Res-CNN-BiLSTM \cite{joshi2022res} for cyberbullying text classification. We provide a detailed comparison of training and testing in results section of this paper respectively.

\section{Methodology}
This section of the paper discusses the various approaches used in this paper. These include a concise explanation of the models used along with their respective parameters.

\subsection{1-D Character Embedding Network}
Character level embedding \cite{zhang2015character} is performed with 1-dimensional convolutional neural network \cite{8682194} that learns the parameters using character level representations. The model starts with embedding layer from input that learns 4830 parameters. Followed by this 1-dimensional convolutional layer is used with 256 filters have 7x7 size each with 3x3 strides which learns 123904 parameters. This uses Rectified Linear Unit abbreviated as ReLU \cite{agarap2018deep} as its activation function. Followed by this 1-dimensional max-pooling layer \cite{wu2015max} is used. Once again then 1-D Convolutional layer is used with same features which learns 459008 parameters. Followed by this is again one more, 1 dimensional max-pooling layer for dimension reduction. This process is repeated 4 times again where the convolutional layers have features of 256 filters with 3x3 dimensions and learn parameters 196864, 196864, 196864 and 196864 respectively. The output from this is flattened for further layers and a dense layer with 1024 hidden neurons, activation function as ReLU with L2 activity and bias regularizers \cite{cortes2012l2} is used. This learns 8913920 parameters. Followed by this one dropout \cite{JMLR:v15:srivastava14a} regularization layer is used with 50\% dropout ratio. This process is repeated again and the dense layer learns 1049600 parameters. The prediction mechanism then uses final 2 layers, one with 32 hidden neurons, activation function as ReLU which learns 32800 parameters followed by final layer of 5 hidden neurons for 5 categories of the output classes with softmax \cite{10.5555/2969830.2969856} activation function that yields probabilities of each class. This layer learns 165 parameters and entire network at the end learns 11,371,683 parameters that roughly translates to 11.3 million parameters. The model uses sparse categorical cross-entropy \cite{zhang2018generalized} loss function along with Adam \cite{kingma2014adam} loss optimizer with learning rate of 3e-4 and decay rate of 5e-6. The batch size of data used is 128 and model runs for 10 epochs respectively.

\subsection{Glove Embedding Network}
The Glove \cite{Pennington2014GloVeGV} embeddings have their respective embeddings in different dimensions. The embedding dimensions we have used are 100 dimensions which is trained on 6 billion corpus of tokens. The input layer is given to 100-dimensional glove embedding that learns 2887000 parameters. Followed by the embeddings the LSTM layer is used with 512 hidden neurons in bidirectional state that learns 2510848 parameters. Following 2 layers are used for final predictions, one dense layer with 32 hidden neurons and ReLU activation function that learns 32800 parameters. The output layer has 5 hidden neurons with softmax \cite{10.5555/2969830.2969856} activation functions that learns 165 parameters. The total parameters learned are 5,430,813 which roughly translates to 5.4 million parameters. The trainable parameters are 2,543,813 and 2,887,000 are non-trainable parameters. Sparse categorical cross-entropy \cite{zhang2018generalized} loss function along with Adam \cite{kingma2014adam} loss optimizer with learning rate of 3e-4 and decay rate of 5e-6 is used. 128 batch size of data for 10 epochs is used respectively.

\subsection{Res-CNN-BiLSTM Network}
This network is combination and modification of 1-D Character Embedding Network and Glove Embedding Network. The Res-CNN-BiLSTM \cite{joshi2022res} is network which trains the 2 networks in parallel fashion and performs concatenation \cite{du2020selective} to yield the final predictions. The steps till 1-dimensional convolutional layer is same as the 1-d character embedding network and the changes in after flattening the network are performed respectively. 4 bidirectional LSTM layers with 512 neurons each are used respectively that return the sequences. The parameters learned are 3149824, 6295552, 6295552, 6295552 respectively. These can actually store a lot of unnecessary information so residuals \cite{he2016deep} are applied to the first and last bidirectional LSTM layer. This although is not sufficient as the major rule to consider for concatenation is the dimensions of independent networks in their final layers should be same. So one more bidirectional LSTM layer with 64 units is used which learns 557568 parameters. The network in total learn 23,969,246 parameters which roughly translates around 23.96 million and are completely trainable. Meanwhile in parallel manner the glove embedding network is trained which uses 100-dimensional 6 billion token glove embedding layer that learns 2887000 parameters followed by 4 bidirectional LSTM layers each consisting of 512 units in returning sequences manner. These layers learn 2510848, 6295552, 6295552 and 6295552 parameters. The first and last layer of bidirectional LSTM are being connected in residual block fashion. So to follow the concatenation rule bidirectional LSTM layer with 64 units is used which learns 557568 parameters. The entire network learns 24,842,072 which is approximately 24.84 million parameters. Now these 2 networks are combined using concatenation process and the rest is used for final prediction with 2 layers. First layer uses 32 hidden neuron dense layer with ReLU \cite{agarap2018deep} activation function followed by output layer of 5 hidden neurons as per 5 output classes with softmax \cite{10.5555/2969830.2969856} activation function. The network uses 2 inputs given in parallel fashion and uses Adam \cite{kingma2014adam} optimizer with sparse categorical cross-entropy \cite{zhang2018generalized} loss function. The entire network learns 48,819,707 which is approximately 48.82 million parameters where trainable parameters are 45,932,707 with 2,887,000 non-trainable parameters.

\subsection{Transformer Based Network}
The gist of the transformer based network lies within the attention \cite{vaswani2017attention} units, whether self-attention or multi-headed-attention \cite{cordonnier2020multi} block. The transformer was originally designed for neural machine translation task but we are using it for text classification task. The attention layer of the transformer is something that separates it from traditional bidirectional based neural networks. It allows the network to take input in entire sequences easily and compute very fast with cost efficient procedures. The attention layer requires input in form of query, key and value which are represented in sequence with entire vector. This attention module is used multiple times hence leading to multiple attention heads. This uses the query, key and value in n-ways parallel manner giving the result independently. The output of all the single attention heads is then merged with a final attention score. This is the basis of a simple transformer block where the attention layers, batch normalization and dropout layers along with a feed-forward neural network are essential parameters. This is not sufficient and requirement of token and positional based embeddings is used. This completes the entire transformer block where the token and position embedding layer is used that learns 643200 parameters. This block is given as an input to transformer block with multi-head attention and learns 10656 parameters. Followed by the transformer block the reduction and usability for traditional layers is done using 1-dimensional global average pooling \cite{lin2013network} layer. The features from the corresponding layer are given to dropout \cite{JMLR:v15:srivastava14a} regularization layer with 50\% threshold. Then a dense layer of 20 hidden neurons is used with ReLU \cite{agarap2018deep} activation function is used that learns 660 parameters. Followed by it one more dropout regularization layer is used. This process is repeated once more and dense layer learns 420 parameters. In the final output layer the dense layer with 5 hidden neurons for 5 output classes is used with softmax \cite{10.5555/2969830.2969856} activation function that learns 105 parameters. The total parameters learned by network are 655,041 which are approximately 600k or 0.65 million. These are way less than expected and inference of these networks will be given in results section in this paper.

\section{Results}
This section of paper gives the inference and effects of every model used and shows how transformer based architecture is able to beat the performance of other used networks in comparison.

\subsection{Training and Validation Accuracy}
\begin{figure}[htbp]
    \centering
    \begin{tabular}{c c}
        \begin{tikzpicture}[scale=0.7]
            \begin{axis}[xlabel={Epochs},ylabel ={Training Accuracy},enlargelimits=false,
            grid=both,
            scale only axis=true,legend pos=south east,style={ultra thick}, axis line style={ultra thin}]
            \addplot+[no markers] table[x=Epochs,y=1-D Char,col sep=comma]{plots/accuracy.csv}; 
            \addplot+[no markers] table[x=Epochs,y=Glove,col sep=comma]{plots/accuracy.csv}; 
            \addplot+[no markers] table[x=Epochs,y=Res-CNN-BiLSTM,col sep=comma]{plots/accuracy.csv};
            \addplot+[no markers] table[x=Epochs,y=Transformer,col sep=comma]{plots/accuracy.csv};
            \addlegendentry{1-D Character Embedding}
            \addlegendentry{Glove Embedding}
            \addlegendentry{Res-CNN-BiLSTM}
            \addlegendentry{Transformer}
            \end{axis}
        \end{tikzpicture}
        &
        \begin{tikzpicture}[scale=0.7]
            \begin{axis}[xlabel = {Epochs},ylabel = {Validation Accuracy},enlargelimits=false,
            grid=both,
            scale only axis=true,legend pos=south east,style={ultra thick}, axis line style={ultra thin}]
            \addplot+[no markers] table[x=Epochs,y=1-D Char,col sep=comma]{plots/validation_accuracy.csv}; 
            \addplot+[no markers] table[x=Epochs,y=Glove,col sep=comma]{plots/validation_accuracy.csv}; 
            \addplot+[no markers] table[x=Epochs,y=Res-CNN-BiLSTM,col sep=comma]{plots/validation_accuracy.csv};
            \addplot+[no markers] table[x=Epochs,y=Transformer,col sep=comma]{plots/validation_accuracy.csv};
            \addlegendentry{1-D Character Embedding}
            \addlegendentry{Glove Embedding}
            \addlegendentry{Res-CNN-BiLSTM}
            \addlegendentry{Transformer}
            \end{axis}
        \end{tikzpicture}
    \end{tabular}
    \caption{Training Accuracy and Validation Accuracy}
    \label{fig:a}
\end{figure}
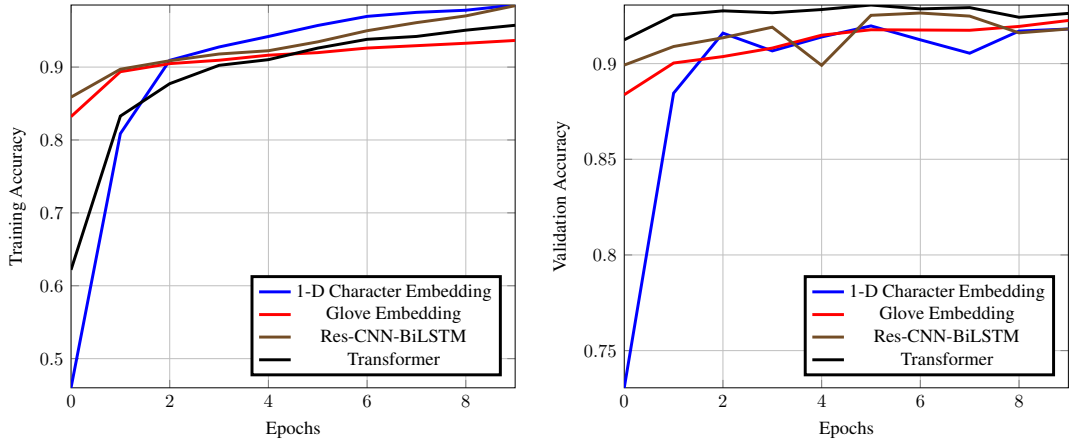

The Accuracy is a commonly used metric for any classification problem and is not just limited to text classification problem. The accuracy is calculated for training and validation for all the respective networks used and can be visualized in the figure \ref{fig:a}. The training accuracy in the last epoch for 1-D Character Embedding is 98.52\% and validation accuracy is 91.81\% respectively. Similarly the training accuracy in the last epoch for Glove Embedding Model is 93.65\% and validation accuracy is 92.25\% respectively. Following the training accuracy for Res-CNN-BiLSTM is 98.42\% and validation accuracy is 91.81\%. Finally the training accuracy for Transformer based network is 95.73\% and validation accuracy is 92.63\%. So practically the generalization of the training and validation accuracy is much better and efficient with transformer based architecture. The accuracy for Res-CNN-BiLSTM in training is higher than Transformer but the validation accuracy for transformer is more. The main noticeable point though for transformer is number of parameters that are around 600k or hardly 0.65 million whereas the parameters for Res-CNN-BiLSTM are around 48.82 million. This states that transformer is still more generalized than Res-CNN-BiLSTM effectively in very minute fraction of parameters. This will give effective outcome in other results in this paper respectively.

\subsection{Training and Validation Loss}
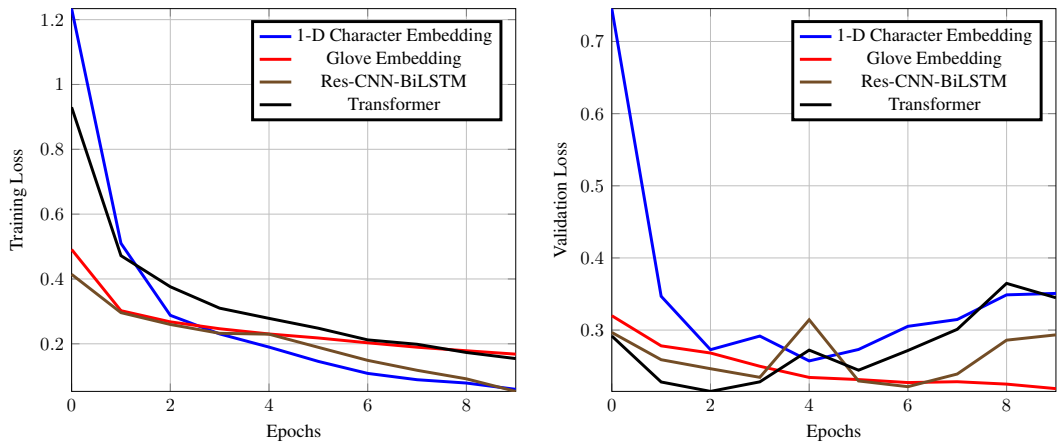
\begin{figure}[htbp]
    \centering
    \begin{tabular}{c c}
        \begin{tikzpicture}[scale=0.7]
            \begin{axis}[xlabel={Epochs},ylabel ={Training Loss},enlargelimits=false,
            grid=both,
            scale only axis=true,legend pos=north east,style={ultra thick}, axis line style={ultra thin}]
            \addplot+[no markers] table[x=Epochs,y=1-D Char,col sep=comma]{plots/loss.csv}; 
            \addplot+[no markers] table[x=Epochs,y=Glove,col sep=comma]{plots/loss.csv}; 
            \addplot+[no markers] table[x=Epochs,y=Res-CNN-BiLSTM,col sep=comma]{plots/loss.csv};
            \addplot+[no markers] table[x=Epochs,y=Transformer,col sep=comma]{plots/loss.csv};
            \addlegendentry{1-D Character Embedding}
            \addlegendentry{Glove Embedding}
            \addlegendentry{Res-CNN-BiLSTM}
            \addlegendentry{Transformer}
            \end{axis}
        \end{tikzpicture}
        &
        \begin{tikzpicture}[scale=0.7]
            \begin{axis}[xlabel = {Epochs},ylabel = {Validation Loss},enlargelimits=false,
            grid=both,
            scale only axis=true,legend pos=north east,style={ultra thick}, axis line style={ultra thin}]
            \addplot+[no markers] table[x=Epochs,y=1-D Char,col sep=comma]{plots/validation_loss.csv}; 
            \addplot+[no markers] table[x=Epochs,y=Glove,col sep=comma]{plots/validation_loss.csv}; 
            \addplot+[no markers] table[x=Epochs,y=Res-CNN-BiLSTM,col sep=comma]{plots/validation_loss.csv};
            \addplot+[no markers] table[x=Epochs,y=Transformer,col sep=comma]{plots/validation_loss.csv};
            \addlegendentry{1-D Character Embedding}
            \addlegendentry{Glove Embedding}
            \addlegendentry{Res-CNN-BiLSTM}
            \addlegendentry{Transformer}
            \end{axis}
        \end{tikzpicture}
    \end{tabular}
    \caption{Training Loss and Validation Loss}
    \label{fig:b}
\end{figure}

The Loss is another indication to what extent there has been a reduction in it. The comparison can be done in a very generalized manner for loss effectively with figure \ref{fig:b}. The training loss for 1-D character embedding 5.59\% whereas the validation loss is 35\% which is not effectively generalized reduction. The training loss for glove embedding model is 16.81\% and validation loss is 21.88\% which is certainly good generalization as compared to 1-D character embedding model. The training loss for Res-CNN-BiLSTM is 5.31\% and validation loss is 29.33\% which is again poor generalization considering the number of parameters involved. Finally the training loss for transformer is 15.43\% and validation loss is 34.47\% which is much better generalized as compared to 1-D character embedding and Res-CNN-BiLSTM but not as efficient as Glove embedding. But consider just 0.65 million parameters for transformer and 5.4 million parameters for Glove, which is almost 9 times, the loss reduction is effectively considerable. Increasing the loss optimization for transformer with adding more multi-head attention can definitely give appropriate performance.

\subsection{Training Time}
The best feature of the transformer is time taken for training and it was one of the primary reasons for which transformer was developed. The time taken by LSTM based architectures were very slower to train. Bidirectional system makes it much slower as the amount of embeddings taken and sentence processed is word based. The transformer leverages this by using an entire sequence of sentence making training very effectively fast. All the networks were trained for 10 epochs under observation and this can be seen on intervals of $1^{st}$, $5^{th}$ and $10^{th}$ epoch respectively in table \ref{tab:a}.

\begin{table}[]
    \caption{Training Time Comparison for Epochs}
    \centering
    \begin{tabular}{llll}
        \toprule
        Algorithm & $1^{st}$ Epoch & $5^{th}$ Epoch & $10^{th}$ Epoch \\
        \midrule
        1-D Char & 29s 96ms & 24s 96ms & 23s 94ms \\
        Glove & 21s 72ms & 17s 71ms & 18s 72ms \\
        Res-CNN-BiLSTM & 359s 690ms & 334s 677ms & 325s 659ms \\
        Transformer & 10s 43ms & 8s 36ms & 6s 29ms \\
        \bottomrule
    \end{tabular}
    \label{tab:a}
\end{table}

From the table \ref{tab:a} it can be inferred that time taken for training the transformer is the fastest. The parameters are one aspect but the nature of the algorithm also makes sense. The time taken by Res-CNN-BiLSTM is highest, then 1-D character embedding model and more faster than it is glove but transformer is almost half times the glove which clearly specifies that using Transformer considering the training time and accuracy trade-off is the best choice.

\section{Conclusion}
Res-CNN-BiLSTM network for Cyberbullying Text Classification had been a good network but had some drawbacks which we wanted to resolve. The use of transformer based architecture was definitely a sure shot move for solving the issue and we did present that in this paper along with results. The transformers have a lot to be explored in near future but proving a small point through a simple application does tingle the curiosity of many researchers. Many advance state-of-the-art transformers have already been developed and used evidently but the main task for us in this in paper was hypothesizing our own custom made Res-CNN-BiLSTM model and we were successful in that. This paper does open many doors to other research with cyberbullying concept and we would like to a part of it with our best belief and understanding.

\section*{Acknowledgments}
We would like to thank Andrew Maranhão for providing the dataset on Kaggle platform. We have also cited the authors from metadata suggested by the author of dataset on Kaggle.

%Bibliography
\bibliographystyle{unsrt}  
\bibliography{references}

\begin{thebibliography}{10}

\bibitem{reshamwala2013review}
Alpa Reshamwala, Dhirendra Mishra, and Prajakta Pawar.
\newblock Review on natural language processing.

\bibitem{young2018recent}
Tom Young, Devamanyu Hazarika, Soujanya Poria, and Erik Cambria.
\newblock Recent trends in deep learning based natural language processing.
\newblock {\em ieee Computational intelligenCe magazine}, 13(3):55--75, 2018.

\bibitem{8629198}
Yasir~Ali Solangi, Zulfiqar~Ali Solangi, Samreen Aarain, Amna Abro, Ghulam~Ali
  Mallah, and Asadullah Shah.
\newblock Review on natural language processing (nlp) and its toolkits for
  opinion mining and sentiment analysis.
\newblock In {\em 2018 IEEE 5th International Conference on Engineering
  Technologies and Applied Sciences (ICETAS)}, pages 1--4, 2018.

\bibitem{lecun2015deep}
Yann LeCun, Yoshua Bengio, and Geoffrey Hinton.
\newblock Deep learning.
\newblock {\em nature}, 521(7553):436--444, 2015.

\bibitem{goodfellow2016deep}
Ian Goodfellow, Yoshua Bengio, and Aaron Courville.
\newblock {\em Deep learning}.
\newblock 2016.

\bibitem{catal2009systematic}
Cagatay Catal and Banu Diri.
\newblock A systematic review of software fault prediction studies.
\newblock {\em Expert systems with applications}, 36(4):7346--7354, 2009.

\bibitem{tan2020neural}
Zhixing Tan, Shuo Wang, Zonghan Yang, Gang Chen, Xuancheng Huang, Maosong Sun,
  and Yang Liu.
\newblock Neural machine translation: A review of methods, resources, and
  tools.
\newblock {\em AI Open}, 1:5--21, 2020.

\bibitem{8629225}
Peng Sun, Xuezhen Yang, Xiaobing Zhao, and Zhijuan Wang.
\newblock An overview of named entity recognition.
\newblock In {\em 2018 International Conference on Asian Language Processing
  (IALP)}, pages 273--278, 2018.

\bibitem{li2020survey}
Jing Li, Aixin Sun, Jianglei Han, and Chenliang Li.
\newblock A survey on deep learning for named entity recognition.
\newblock {\em IEEE Transactions on Knowledge and Data Engineering},
  34(1):50--70, 2020.

\bibitem{roy2021recent}
Arya Roy.
\newblock Recent trends in named entity recognition (ner).
\newblock {\em arXiv preprint arXiv:2101.11420}, 2021.

\bibitem{sah2020machine}
Shagan Sah.
\newblock Machine learning: a review of learning types.
\newblock 2020.

\bibitem{dhall2020machine}
Devanshi Dhall, Ravinder Kaur, and Mamta Juneja.
\newblock Machine learning: a review of the algorithms and its applications.
\newblock {\em Proceedings of ICRIC 2019}, pages 47--63, 2020.

\bibitem{naseem2021comprehensive}
Usman Naseem, Imran Razzak, Shah~Khalid Khan, and Mukesh Prasad.
\newblock A comprehensive survey on word representation models: From classical
  to state-of-the-art word representation language models.
\newblock {\em Transactions on Asian and Low-Resource Language Information
  Processing}, 20(5):1--35, 2021.

\bibitem{zhang2010understanding}
Yin Zhang, Rong Jin, and Zhi-Hua Zhou.
\newblock Understanding bag-of-words model: a statistical framework.
\newblock {\em International Journal of Machine Learning and Cybernetics},
  1(1):43--52, 2010.

\bibitem{kim2019research}
Sang-Woon Kim and Joon-Min Gil.
\newblock Research paper classification systems based on tf-idf and lda
  schemes.
\newblock {\em Human-centric Computing and Information Sciences}, 9(1):1--21,
  2019.

\bibitem{Ramos2003UsingTT}
Juan~Enrique Ramos.
\newblock Using tf-idf to determine word relevance in document queries.
\newblock 2003.

\bibitem{tambe2022effects}
Sayali Tambe, Raunak Joshi, Abhishek Gupta, Nandan Kanvinde, and Vidya Chitre.
\newblock Effects of parametric and non-parametric methods on high dimensional
  sparse matrix representations.
\newblock {\em arXiv preprint arXiv:2202.02894}, 2022.

\bibitem{almeida2019word}
Felipe Almeida and Geraldo Xex{\'e}o.
\newblock Word embeddings: A survey.
\newblock {\em arXiv preprint arXiv:1901.09069}, 2019.

\bibitem{mikolov2013efficient}
Tomas Mikolov, Kai Chen, Greg Corrado, and Jeffrey Dean.
\newblock Efficient estimation of word representations in vector space.
\newblock {\em arXiv preprint arXiv:1301.3781}, 2013.

\bibitem{cerda2018similarity}
Patricio Cerda, Ga{\"e}l Varoquaux, and Bal{\'a}zs K{\'e}gl.
\newblock Similarity encoding for learning with dirty categorical variables.
\newblock {\em Machine Learning}, 107(8):1477--1494, 2018.

\bibitem{Pennington2014GloVeGV}
Jeffrey Pennington, Richard Socher, and Christopher~D. Manning.
\newblock Glove: Global vectors for word representation.
\newblock In {\em EMNLP}, 2014.

\bibitem{Landauer1998AnIT}
Thomas~K. Landauer, Peter~W. Foltz, and Darrell Laham.
\newblock An introduction to latent semantic analysis.
\newblock {\em Discourse Processes}, 25:259--284, 1998.

\bibitem{Mahmood2013LiteratureSO}
A.~S. M.~Ashique Mahmood.
\newblock Literature survey on topic modeling.
\newblock 2013.

\bibitem{zhang2015character}
Xiang Zhang, Junbo Zhao, and Yann LeCun.
\newblock Character-level convolutional networks for text classification.
\newblock {\em Advances in neural information processing systems}, 28, 2015.

\bibitem{8682194}
Serkan Kiranyaz, Turker Ince, Osama Abdeljaber, Onur Avci, and Moncef Gabbouj.
\newblock 1-d convolutional neural networks for signal processing applications.
\newblock In {\em ICASSP 2019 - 2019 IEEE International Conference on
  Acoustics, Speech and Signal Processing (ICASSP)}, pages 8360--8364, 2019.

\bibitem{Cramer2002TheOO}
J.~S. Cramer.
\newblock The origins of logistic regression.
\newblock {\em Econometrics eJournal}, 2002.

\bibitem{708428}
M.A. Hearst, S.T. Dumais, E.~Osuna, J.~Platt, and B.~Scholkopf.
\newblock Support vector machines.
\newblock {\em IEEE Intelligent Systems and their Applications}, 13(4):18--28,
  1998.

\bibitem{zhang2012support}
Yongli Zhang.
\newblock Support vector machine classification algorithm and its application.
\newblock In {\em International conference on information computing and
  applications}, pages 179--186. Springer, 2012.

\bibitem{cunningham2021k}
Padraig Cunningham and Sarah~Jane Delany.
\newblock k-nearest neighbour classifiers-a tutorial.
\newblock {\em ACM Computing Surveys (CSUR)}, 54(6):1--25, 2021.

\bibitem{breiman1996bagging}
Leo Breiman.
\newblock Bagging predictors.
\newblock {\em Machine learning}, 24(2):123--140, 1996.

\bibitem{breiman2001random}
Leo Breiman.
\newblock Random forests.
\newblock {\em Machine learning}, 45(1):5--32, 2001.

\bibitem{kanvinde2022binary}
Nandan Kanvinde, Abhishek Gupta, and Raunak Joshi.
\newblock Binary classification for high dimensional data using supervised
  non-parametric ensemble method.
\newblock {\em arXiv preprint arXiv:2202.07779}, 2022.

\bibitem{freund1996experiments}
Yoav Freund, Robert~E Schapire, et~al.
\newblock Experiments with a new boosting algorithm.
\newblock In {\em icml}, volume~96, pages 148--156. Citeseer, 1996.

\bibitem{10.5555/1624312.1624417}
Robert~E. Schapire.
\newblock A brief introduction to boosting.
\newblock In {\em Proceedings of the 16th International Joint Conference on
  Artificial Intelligence - Volume 2}, IJCAI'99, page 1401–1406, San
  Francisco, CA, USA, 1999. Morgan Kaufmann Publishers Inc.

\bibitem{schapire2003boosting}
Robert~E Schapire.
\newblock The boosting approach to machine learning: An overview.
\newblock {\em Nonlinear estimation and classification}, pages 149--171, 2003.

\bibitem{gupta2021succinct}
Abhishek~M Gupta, Sannidhi~S Shetty, Raunak~M Joshi, and Ronald~Melwin Laban.
\newblock Succinct differentiation of disparate boosting ensemble learning
  methods for prognostication of polycystic ovary syndrome diagnosis.
\newblock In {\em 2021 International Conference on Advances in Computing,
  Communication, and Control (ICAC3)}, pages 1--5. IEEE, 2021.

\bibitem{Ramayah2010DiscriminantA}
Thurasamy Ramayah, Noor~Hazlina Ahmad, Hasliza~Abdul Halim, Siti
  Rohaida~Mohamed Zainal, and May-Chiun Lo.
\newblock Discriminant analysis : An illustrated example.
\newblock {\em African Journal of Business Management}, 4:1654--1667, 2010.

\bibitem{gupta2022discriminant}
Abhishek Gupta, Himanshu Soni, Raunak Joshi, and Ronald~Melwin Laban.
\newblock Discriminant analysis in contrasting dimensions for polycystic ovary
  syndrome prognostication.
\newblock {\em arXiv preprint arXiv:2201.03029}, 2022.

\bibitem{Wolpert1992StackedG}
David~H. Wolpert.
\newblock Stacked generalization.
\newblock {\em Neural Networks}, 5:241--259, 1992.

\bibitem{nair2022combining}
Sruthi Nair, Abhishek Gupta, Raunak Joshi, and Vidya Chitre.
\newblock Combining varied learners for binary classification using stacked
  generalization.
\newblock {\em arXiv preprint arXiv:2202.08910}, 2022.

\bibitem{cho2014learning}
Kyunghyun Cho, Bart Van~Merri{\"e}nboer, Caglar Gulcehre, Dzmitry Bahdanau,
  Fethi Bougares, Holger Schwenk, and Yoshua Bengio.
\newblock Learning phrase representations using rnn encoder-decoder for
  statistical machine translation.
\newblock {\em arXiv preprint arXiv:1406.1078}, 2014.

\bibitem{10.1142/S0218488598000094}
Sepp Hochreiter.
\newblock The vanishing gradient problem during learning recurrent neural nets
  and problem solutions.
\newblock {\em Int. J. Uncertain. Fuzziness Knowl.-Based Syst.},
  6(2):107–116, apr 1998.

\bibitem{rumelhart1986learning}
David~E Rumelhart, Geoffrey~E Hinton, and Ronald~J Williams.
\newblock Learning representations by back-propagating errors.
\newblock {\em nature}, 323(6088):533--536, 1986.

\bibitem{hochreiter1997long}
Sepp Hochreiter and J{\"u}rgen Schmidhuber.
\newblock Long short-term memory.
\newblock {\em Neural computation}, 9(8):1735--1780, 1997.

\bibitem{olah2015understanding}
Christopher Olah.
\newblock Understanding lstm networks.
\newblock 2015.

\bibitem{staudemeyer2019understanding}
Ralf~C Staudemeyer and Eric~Rothstein Morris.
\newblock Understanding lstm--a tutorial into long short-term memory recurrent
  neural networks.
\newblock {\em arXiv preprint arXiv:1909.09586}, 2019.

\bibitem{650093}
M.~Schuster and K.K. Paliwal.
\newblock Bidirectional recurrent neural networks.
\newblock {\em IEEE Transactions on Signal Processing}, 45(11):2673--2681,
  1997.

\bibitem{vaswani2017attention}
Ashish Vaswani, Noam Shazeer, Niki Parmar, Jakob Uszkoreit, Llion Jones,
  Aidan~N Gomez, {\L}ukasz Kaiser, and Illia Polosukhin.
\newblock Attention is all you need.
\newblock {\em Advances in neural information processing systems}, 30, 2017.

\bibitem{9378065}
Jason Wang, Kaiqun Fu, and Chang-Tien Lu.
\newblock Sosnet: A graph convolutional network approach to fine-grained
  cyberbullying detection.
\newblock In {\em 2020 IEEE International Conference on Big Data (Big Data)},
  pages 1699--1708, 2020.

\bibitem{joshi2022res}
Raunak Joshi, Abhishek Gupta, and Nandan Kanvinde.
\newblock Res-cnn-bilstm network for overcoming mental health disturbances
  caused due to cyberbullying through social media.
\newblock {\em arXiv preprint arXiv:2204.09738}, 2022.

\bibitem{agarap2018deep}
Abien~Fred Agarap.
\newblock Deep learning using rectified linear units (relu).
\newblock {\em arXiv preprint arXiv:1803.08375}, 2018.

\bibitem{wu2015max}
Haibing Wu and Xiaodong Gu.
\newblock Max-pooling dropout for regularization of convolutional neural
  networks.
\newblock In {\em International Conference on Neural Information Processing},
  pages 46--54. Springer, 2015.

\bibitem{cortes2012l2}
Corinna Cortes, Mehryar Mohri, and Afshin Rostamizadeh.
\newblock L2 regularization for learning kernels.
\newblock {\em arXiv preprint arXiv:1205.2653}, 2012.

\bibitem{JMLR:v15:srivastava14a}
Nitish Srivastava, Geoffrey Hinton, Alex Krizhevsky, Ilya Sutskever, and Ruslan
  Salakhutdinov.
\newblock Dropout: A simple way to prevent neural networks from overfitting.
\newblock {\em Journal of Machine Learning Research}, 15(56):1929--1958, 2014.

\bibitem{10.5555/2969830.2969856}
John~S. Bridle.
\newblock Training stochastic model recognition algorithms as networks can lead
  to maximum mutual information estimation of parameters.
\newblock In {\em Proceedings of the 2nd International Conference on Neural
  Information Processing Systems}, NIPS'89, page 211–217, Cambridge, MA, USA,
  1989. MIT Press.

\bibitem{zhang2018generalized}
Zhilu Zhang and Mert Sabuncu.
\newblock Generalized cross entropy loss for training deep neural networks with
  noisy labels.
\newblock {\em Advances in neural information processing systems}, 31, 2018.

\bibitem{kingma2014adam}
Diederik~P Kingma and Jimmy Ba.
\newblock Adam: A method for stochastic optimization.
\newblock {\em arXiv preprint arXiv:1412.6980}, 2014.

\bibitem{du2020selective}
Chen Du, Yanna Wang, Chunheng Wang, Cunzhao Shi, and Baihua Xiao.
\newblock Selective feature connection mechanism: Concatenating multi-layer cnn
  features with a feature selector.
\newblock {\em Pattern Recognition Letters}, 129:108--114, 2020.

\bibitem{he2016deep}
Kaiming He, Xiangyu Zhang, Shaoqing Ren, and Jian Sun.
\newblock Deep residual learning for image recognition.
\newblock In {\em Proceedings of the IEEE conference on computer vision and
  pattern recognition}, pages 770--778, 2016.

\bibitem{cordonnier2020multi}
Jean-Baptiste Cordonnier, Andreas Loukas, and Martin Jaggi.
\newblock Multi-head attention: Collaborate instead of concatenate.
\newblock {\em arXiv preprint arXiv:2006.16362}, 2020.

\bibitem{lin2013network}
Min Lin, Qiang Chen, and Shuicheng Yan.
\newblock Network in network.
\newblock {\em arXiv preprint arXiv:1312.4400}, 2013.

\end{thebibliography}

\end{document}